\documentclass{vgtc}                          

\ifpdf
  \pdfoutput=1\relax                   
  \usepackage{graphicx}                
  \DeclareGraphicsExtensions{.pdf,.png,.jpg,.jpeg} 
\else
  \usepackage{graphicx}                
  \DeclareGraphicsExtensions{.eps}     
\fi%

\graphicspath{{figures/}{pictures/}{images/}{./}} 

\usepackage{microtype}                 
\PassOptionsToPackage{warn}{textcomp}  
\usepackage{textcomp}                  
\usepackage{mathptmx}                  
\usepackage{times}                     
\usepackage{cite}                      
\usepackage{tabu}                      
\usepackage{booktabs}                  
\usepackage{bbm}
\usepackage{color}
\usepackage{booktabs}
\usepackage{amsmath}
\usepackage{bm}
\usepackage{algorithm}
\usepackage{algorithmic}

\onlineid{0}

\vgtccategory{Research}

\vgtcinsertpkg

\title{Adaptive Local Adversarial Attacks on 3D Point Clouds for Augmented Reality}

\author{Weiquan Liu\thanks{e-mail: wqliu@xmu.edu.cn}
\and Shijun Zheng\thanks{e-mail: zhengshijun@stu.xmu.edu.cn}
\and Cheng Wang\thanks{Corresponding author, e-mail: cwang@xmu.edu.cn}}
\affiliation{Fujian Key Laboratory of Sensing and Computing for Smart Cities, School of Informatics, \\Xiamen University, Xiamen, China}

\teaser{
  \centering
  \includegraphics[width=\linewidth]{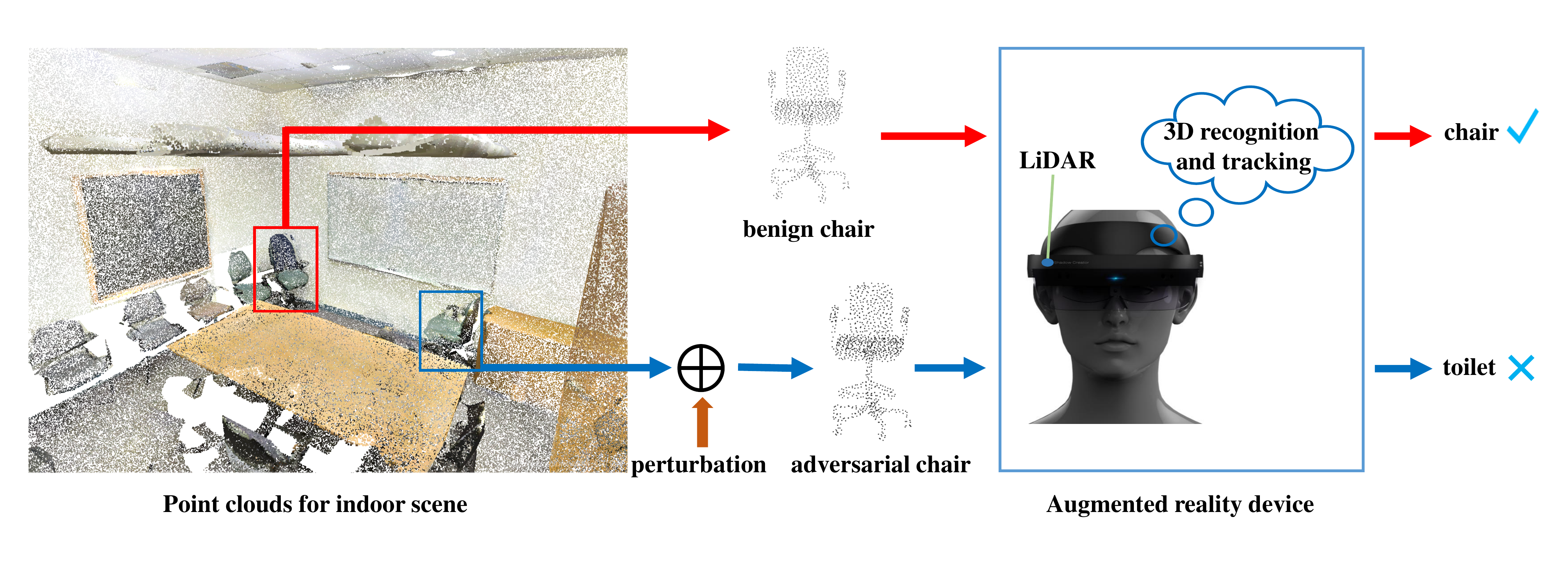}
  \caption{Adversarial examples pose security risks to Augmented Reality systems. In indoor scenes, AR devices use LiDAR to perceive the scene, and apply 3D recognition and tracking technology to recognize  and track objects in real time. Benign objects are correctly detected and recognized by AR devices, for example, a benign "chair" is correctly recognize. Adding perturbations to benign objects to generate adversarial objects that AR devices cannot correctly recognize. For example, an adversarial "chair" is recognized by the AR device as a "toilet".}
  \label{fig:teaser}
}

\abstract{As the key technology of augmented reality (AR), 3D recognition and tracking are always vulnerable to adversarial examples, which will cause serious security risks to AR systems. Adversarial examples are beneficial to improve the robustness of the 3D neural network model and enhance the stability of the AR system. At present, most 3D adversarial attack methods perturb the entire point cloud to generate adversarial examples, which results in high perturbation costs and difficulty in reconstructing the corresponding real objects in the physical world. In this paper, we propose an adaptive local adversarial attack method (AL-Adv) on 3D point clouds to generate adversarial point clouds. First, we analyze the vulnerability of the 3D network model and extract the salient regions of the input point cloud, namely the vulnerable regions. Second, we propose an adaptive gradient attack algorithm that targets vulnerable regions. The proposed attack algorithm adaptively assigns different disturbances in different directions of the three-dimensional coordinates of the point cloud. Experimental results show that our proposed method AL-Adv achieves a higher attack success rate than the global attack method. Specifically, the adversarial examples generated by the AL-Adv demonstrate good imperceptibility and small generation costs.
} 

\CCScatlist{
  \CCScatTwelve{augmented reality}{3D adversarial attack}{point cloud}{local regions attack};
}

\begin{document}


\firstsection{Introduction}

\maketitle

In recent years, with the development of deep learning technology, AR technology has been deeply combined with deep learning, such as 2D and 3D object detection, recognition, and so on \cite{che2021detection,billinghurst2015survey,de2020survey}. In AR, determining the three-dimensional (3D) spatial position of objects in the real world is an important task, which directly affects the user's product experience. To ensure the better superposition of virtual objects on real objects, AR systems have higher requirements for real-time recognition and tracking of objects. AR object recognition and tracking based on deep learning methods (3D deep network) \cite{kastner20203d,han2020live}, a core technology of augmented reality, has been widely used in the industrial field, providing an intuitive and efficient solution for remote guidance, maintenance, and training \cite{gattullo2020and}. However, there is still a great security threat in the actual application process of AR object recognition and tracking based on 3D deep network. If the attacker produces adversarial objects in the real world, it will largely cause the AR system to fail to perceive the real scene normally \cite{huang2022spaa}.

The recognition and tracking of the 3D point cloud \cite{deng2022vista,zheng2022beyond} is an important support for the application of AR. AR realizes the perception of the scene by detecting, recognizing, and tracking objects in the real environment, which is the basis for the seamless overlay of virtual objects. However, previous studies \cite{tu2020physically,xiang2019generating,liu2021pointguard,zhao2020isometry,sun2021local} found that 3D deep network models are vulnerable to adversarial examples, resulting in the models producing erroneous recognition results. For the input 3D point cloud, adversarial examples are generated by adversarial perturbations, which cause the 3D deep network model to predict wrong results. If the 3D network model is attacked by adversarial examples, the 3D recognition and tracking functions of the AR system will not work properly, which is a great risk for AR applications. In 3D point clouds, adversarial attacks are mainly divided into two categories: the digital domain and the physical world. 

In the digital domain, adversarial examples are mainly generated by adding points or clusters, removing points, and perturbing points \cite{xiang2019generating,zheng2019pointcloud}. The generated adversarial examples are required to be as close as possible to the original point cloud while maintaining good 3D properties such as smoothness and fairness. At the same time, the adversarial examples should have a high attack success rate on the 3D network model.

In the physical world, 3D adversarial examples need to be constructed in real scenes. For example, adversarial examples generated in the digital are presented in the real world through 3D printing \cite{wen2020geometry,cao2021invisible}, and they are used to attack artificial intelligence systems based on 3D network models. Such reconstructed adversarial examples in the physical world often pose a greater threat to artificial intelligence systems, including AR systems. Therefore, this paper mainly studies how to generate adversarial examples of 3D point clouds, which are crucial for the robustness of 3D network models. The stable recognition and tracking technology of the 3D network model helps to improve the robustness of the AR system.

In general 3D network models, the training and testing samples are often benign, which also leads to the model always making mistakes in recognizing adversarial examples. Therefore, researchers usually add adversarial examples to the training set, and the model learns them while training. This approach can effectively deal with the threat of adversarial examples to the 3D network model \cite{ren2022benchmarking}. We mainly focus on how to generate high-quality 3D point cloud adversarial examples, which are beneficial to improve the security of AR systems. In summary, the better the imperceptibility of the generated adversarial examples, the harder it is for the human eye to perceive.

In this paper, we pay attention to local regions of 3D point clouds rather than the whole point cloud when generating adversarial examples. Compared to attacking the entire point cloud, adversarial examples generated by attacking local regions of the point cloud have better operability in the physical world. Therefore, this paper proposes an adaptive local adversarial attack method (AL-Adv) to generate high-quality adversarial point clouds. First, we introduce the idea of game theory to analyze the 3D network model. Specifically, Shapley value is used to analyze the vulnerability of 3D network models and extract salient regions. Each region uses a Shapley value to represent its importance to the recognition results of the network model. If the saliency of a region is stronger, the 3D network model is more vulnerable in this region. Second, we design an adaptive gradient attack algorithm for salient regions. The attack algorithm adaptively assigns different perturbations to each direction of the 3D coordinates according to the gradient of the point cloud. To verify the effectiveness of the generated adversarial examples, the proposed AL-Adv is compared with several popular adversarial attack methods. Experimental results show that the proposed AL-Adv has a higher attack success rate than other global adversarial attack methods.

The main contributions of this work can be expressed as follows:
\begin{itemize}
\item Existing adversarial attack methods mainly focus on the global point cloud, while our proposed AL-Adv method pays more attention to the local regions of the point cloud.
\item To obtain higher-quality 3D adversarial point clouds, we design a novel adaptive gradient attack algorithm for local regions.
\item The proposed method AL-Adv achieves a higher attack success rate in local regions compared to existing global adversarial attack methods.
\end{itemize}

\section{Related works}
At present, the object recognition and tracking methods used in AR systems are mainly divided into two categories: 2D recognition and 3D recognition. This paper mainly focuses on the impact of 3D point cloud object recognition and tracking methods on AR systems. Therefore, this section presents the related works of adversarial attacks on 3D point clouds, both in the digital domain and in the physical world.

\subsection{Digital adversarial attack}
In images, samples obtained by applying adversarial perturbations to the input image cause the deep network model to produce incorrect predictions, such samples are called adversarial examples\cite{szegedy2013intriguing}. For images, many methods of generating adversarial examples have emerged \cite{goodfellow2014explaining,kurakin2018adversarial,zheng2019distributionally,carlini2017towards,dong2018boosting,papernot2016limitations,xiao2018generating,athalye2018obfuscated}. When generating adversarial examples, the goal is to achieve a high attack success rate while making them imperceptible to the human eye. Distance constraints are usually used in this process to limit the generation of adversarial examples. In 3D point clouds, many adversarial attack methods are extended from images \cite{liu2019extending,xiang2019generating}. The most common way to generate 3D point cloud adversarial examples is to perturb, remove, and add points to the input point cloud. For example, Xiang et al.\cite{xiang2019generating} perturbed the entire point cloud so that the points of the point cloud deviate from the original position to deceive the 3D network model. In addition, point clusters or objects with shapes (such as planes, balls, etc.) are added to the input point cloud to generate adversarial point clouds. Zheng et al. \cite{zheng2019pointcloud} proposed a new adversarial attack method by removing points from the input point cloud to generate adversarial examples. First, the method generated a saliency map of the input point cloud through the gradient, which represented the contribution of each point to the model prediction result. Therefore, the corresponding point removal algorithm was designed for the highly significant points to generate adversarial examples effectively. As the number of removed points increases, the recognition results of the 3D network model were decreased. To ensure the correct working of the 3D network model, the operation of point removal was equivalent to moving the point to the centroid of the point cloud. To enhance the transferability of adversarial examples and make them more difficult to defend, Hamdi et al. \cite{hamdi2020advpc} proposed a data-driven adversarial attack method. This method designed a new loss function to perturb the input point cloud.

Using an optimized method to generate adversarial examples is another important way by designing the objective function of adversarial attacks. Kim et al. \cite{kim2021minimal} proposed a unified adversarial point cloud generation formulation to obtain better imperceptible adversarial point clouds with minimal manipulation points. This method unifies the two adversarial point cloud generation methods of perturbed points and added points, and used distance constraint and point number constraint when generating adversarial examples. To avoid obvious outliers in adversarial examples and maintain the basic properties of 3D objects, Wen et al. \cite{wen2020geometry} designed a novel geometry-aware objective function to generate adversarial point clouds. Then based on this geometry-aware objective function, an optimization-based adversarial attack method was implemented by regularizing the adversarial loss. The generated adversarial point cloud was more harmful to the 3D network model and harder to defend. However, Huang et al. \cite{huang2022shape} argued that 3D point clouds were highly structured data, and it was difficult to use simple constraints to limit perturbations. Therefore, the method transformed the original point cloud into a new coordinate system and constrains the movement of points within the tangent plane. In addition, the method used gradients in the new coordinate system to find the best attack direction and built a saliency map of the point cloud.

\begin{figure*}[!tb]
	\centering 
	\includegraphics[width= \linewidth]{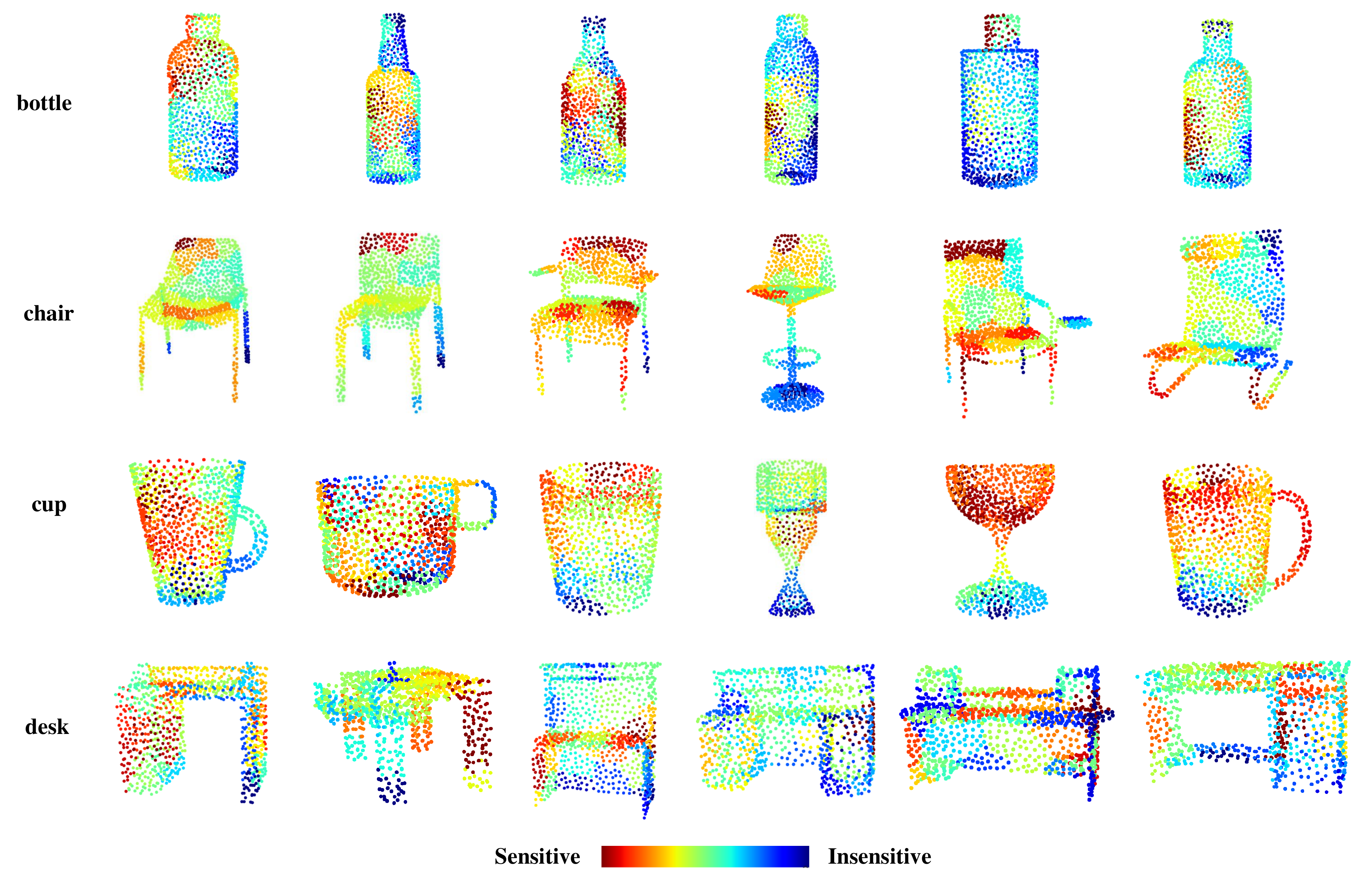}
	\caption{The salient regions of the point clouds. A vulnerability analysis of PointNet was performed on ModelNet40.}
	\label{fig:saliencymap}
\end{figure*}

\subsection{Physical adversarial attack}
Currently, for the 3D point cloud, the digital adversarial attack is the main research direction, and there is less work on the adversarial attack in the physical world. There are three categories of existing adversarial attacks in the physical world. The first category is to generate high-quality adversarial examples with a high attack success rate through digital simulation, and then use 3D printing technology to reconstruct the adversarial object in the real world. For example, Wen et al. \cite{wen2020geometry} first used an adversarial attack method to generate adversarial point clouds and convert the adversarial point clouds into meshes. The reconstructed adversarial meshes were then produced in the real world using 3D printing. Finally, the printed real objects were rescanned into point clouds for testing the performance of the 3D network model. The second category is to implement adversarial attacks in point cloud data scanned in real scenes. For example, Tu et al. \cite{tu2020physically} proposed an adversarial attack method to generate adversarial objects of different geometries. Placing these adversarial objects on top of the car's point cloud can fool the LiDAR, causing the 3D object recognition network to fail to detect the car. The third category is the realization of adversarial attacks in the physical world, leading to the failure of LiDAR. Zhu et al. \cite{zhu2021can} proposed an adversarial attack framework for finding attack locations in the real world. Placing any object with a reflective surface, such as a commercial drone, at these attack positions makes the target object invisible to the LiDAR. This attack method posed a very large threat to autonomous driving systems in the real world.

\begin{figure*}[!tb]
	\centering 
	\includegraphics[width=0.8 \linewidth]{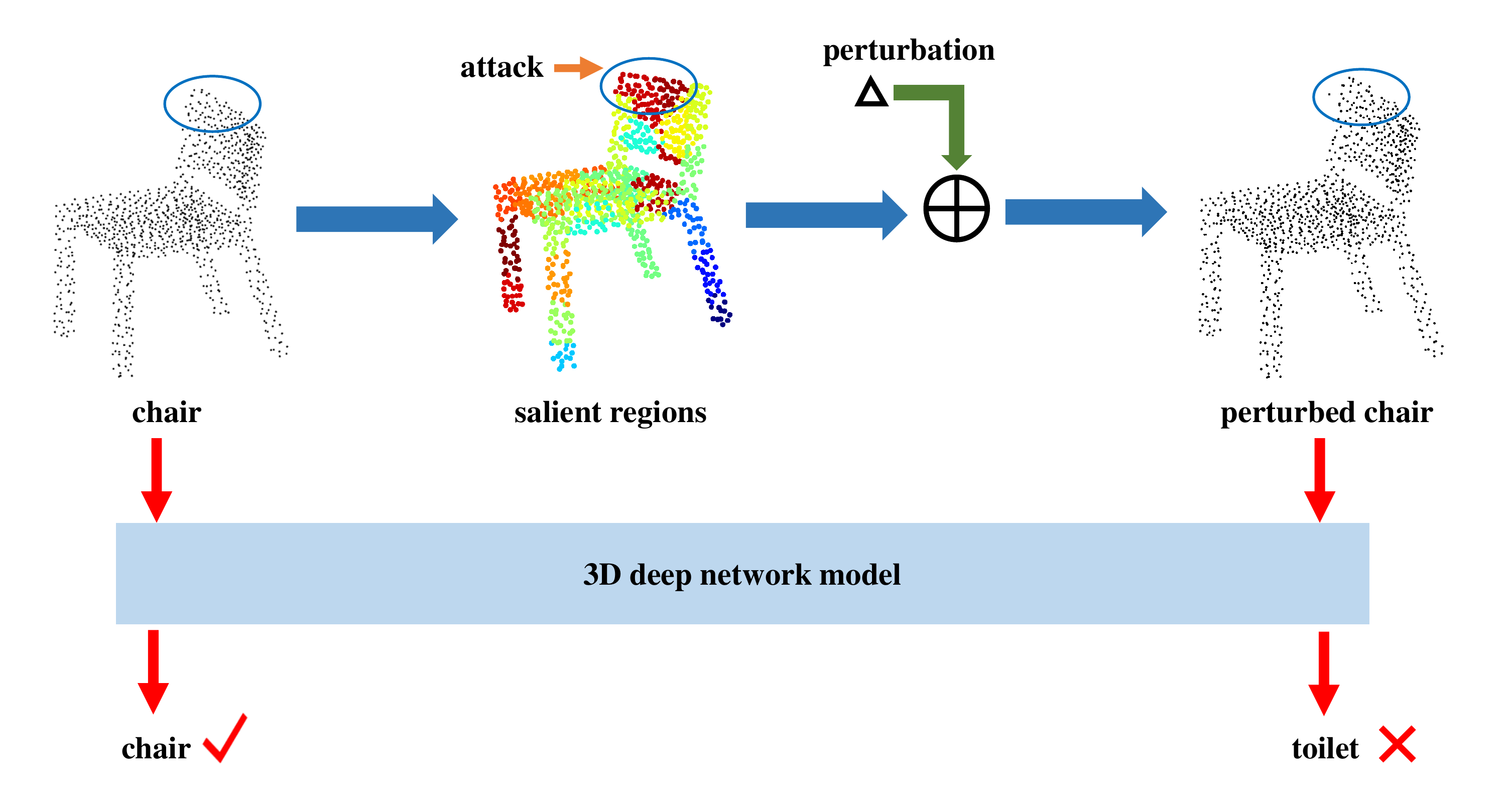}
	\caption{The point cloud attack scheme based on salient regions. Perturbations are added to salient regions of benign point cloud "chair" to generate adversarial point cloud "perturbed chair". The benign point cloud "chair" is correctly identified, and the adversarial point cloud "perturbed chair" is identified as "toilet".}
	\label{fig:attackpiple}
\end{figure*}

\section{Method}
\subsection{Vulnerability Analysis of 3D Network Models}
\subsubsection{Shapley value}
We are more concerned with the effect of local regions of the point cloud on the output of the network model than the global structure of the point cloud. We introduce how the Shapley value is applied to the vulnerability analysis of 3D network models. The Shapley value is a reasonable method used in game theory \cite{shapley1997value} to address the distribution of cooperative benefits. Supposing there are multiple players participating in the game, and different players will get different rewards. In the whole game process, multiple players cooperate to obtain the maximum reward, some players contribute greatly to the final reward, and some players contribute less. The Shapley value is used to reasonably distribute rewards to each player to ensure fairness.

\subsubsection{Shapley value for 3D network models}
In this paper, the Shapley value is used to analyze the vulnerability of 3D network models. Specifically, we take the 3D network model as the game, divide the input point cloud into $m$ regions, that is, $m$ players, and the output of the 3D network model is used as a reward. We use the Shapley value to fairly assign rewards to these $m$ regions. If a region plays a more important role in the output of the network model, it is assigned a higher reward. Conversely, if a region contributes less to the output of the network model, then it also receives less reward.

For an input point cloud $x$, we divide it into $m$ regions, denoted as $x=(a_1,a_2,\cdot\cdot\cdot,a_k,\cdot\cdot\cdot,a_m)$, and $a_i$ represents the $i$-th region. Then for the input point cloud $x$, the set of all players is represented as $M=\{1, 2, ..., m\}$. Given a trained 3D network model, its output is denoted as $g(\cdot)$. Assuming that $S\in M$ is a set of some of these players, the rewards obtained by players in the set $S$ when they participate in the game are denoted as $g(S)$. Therefore, the Shapley value of the $i$-th region of the input point cloud is represented as $\phi(i)$.
\begin{equation}
	\phi(i)=\sum_{S \subseteq M \setminus \{i\}} \frac{|S| !(m-|S|-1) !}{m !}(g(S \cup\{i\})-g(S)).
\end{equation}

In detail, $\phi(i)$ represents the importance of the $i$-th region of the input point cloud to the recognition result of the 3D network model. The higher the value of $\phi(i)$, the more important the $i$-th region is to the recognition result of the network model. Therefore, each region is assigned a salient value using the Shapley value, which indicates how important the region is to the output of the network model. That is to say, the higher the salient value of a region, the greater the effect of the region on the recognition results of the model. For 3D network models, regions with higher salient values are more vulnerable.

The Shapley value is used to analyze the vulnerability of the 3D point cloud deep network model and extract the salient regions of the point cloud, as shown in \autoref{fig:saliencymap}.

\subsection{Generating Adversarial Point Clouds}
\textbf{Untargeted Adversarial Attacks.} In this paper, we focus on untargeted attacks on 3D network models. For a point cloud $x \in R^{(n \times 3)} $containing $n$ points, its ground-truth label is $y$. The 3D network model $f$ can correctly identify $x$, that is, $f(x)=y$. Our goal is to find a point cloud $x'$, which makes the recognition result of the 3D network model $f$ wrong, that is, $f(x')\neq y$. Furthermore, in the process of generating $x'$, it should be ensured that x and $x'$ are as similar as possible. Such $x'$ is called an adversarial example and can be expressed as follows:

\begin{equation}
	\min D\left(x, x^{\prime}\right) \quad \text { s.t. } f\left(x^{\prime}\right)\neq y ,
\end{equation}
\noindent where $D$ represents a disturbance metric between $x$ and $x'$, such as a distance metric. The generated adversarial point cloud is forced to be closer to the original point cloud by minimizing the metric $D$. A smaller value of the metric $D$ indicates that the adversarial point cloud $x'$ is closer to the original point cloud $x$ and more difficult to distinguish.

\textbf{Adversarial Attacks on Local Regions.}
We have used Shapley value to perform a vulnerability analysis on a 3D network model and extracted vulnerable regions. Our goal is to only attack a certain number of vulnerable regions to generate adversarial examples, not to attack the entire point cloud. The point cloud attack scheme based on salient regions is shown in \autoref{fig:attackpiple}.

Most adversarial attack methods are to perturb the entire point cloud, however, we are more concerned with the local regions of the point cloud. Therefore, we divide the original point cloud into $m$ regions, denoted as $x=(a_1, a_2,\cdot\cdot\cdot,a_k,\cdot\cdot\cdot,a_m)$, $a_i$ representing the $i$-th region. Assuming that the perturbation of the region $a_i$ is $e_i$, then the adversarial point cloud $x'$ can be expressed as follows:
\begin{equation}
	x'= \{a'_i=a_i+e_i | a_i \in x \} .
\end{equation}

It can be seen that $x$ and $x'$ have the same structure, that is, the adversarial point cloud $x'=(a_1',a_2',\cdot\cdot\cdot,a_i',\cdot\cdot\cdot,a_m')$, where $a_i'$ is obtained by perturbing the region $a_i$.

Therefore, according to Equation (2), the regional attack of 3D point cloud can be formulated as follows:

\begin{equation}
	\min C_{Region}= l(x') + \lambda_{1} \ast D(x, x') + \lambda_{2}\ast P(x, x').
\end{equation}

$l(x')$ is the adversarial loss, $D(x,x')$ is the distance constraint between the original point cloud $x$ and the adversarial point cloud $x'$. $\lambda_{1}$ and $\lambda_{2}$ are constants. In the actual solution process, the binary search method is used to automatically find the optimal parameter value of $\lambda_{1}$. $P(x, x')$ represents the number of points modified by the adversarial point cloud compared to the original point cloud. Finally, adversarial examples are generated by solving equation (4).

\textbf{Perceptibility.}
When generating adversarial examples, it is necessary to use relevant constraints to ensure that the adversarial examples are as close as possible to the original examples. Using distance constraints is an effective way to improve the imperceptibility of generated adversarial examples. This paper uses the Chamfer distance and Hausdorff distance, which are common in 3D data, to constrain the generation of adversarial point clouds. These two distance constraints are also important indicators to measure the quality of adversarial point clouds.

Chamfer distance is used to measure the difference between two point sets. The calculation is as follows:

\begin{equation}
    \begin{aligned}
        D_{Chamfer}\left(x, x^{\prime}\right)=\max \left\{\frac{1}{n} \sum_{b \in x} \min _{a \in x^{\prime}}\|b-a\|_2^2,\right. \\
    \left.\frac{1}{n^{\prime}} \sum_{a \in x^{\prime}} \min _{b \in x}\|a-b\|_2^2\right\},
    \end{aligned}
\end{equation}

\noindent where $n$ represents the number of points of the original point cloud $x$, and $n'$ represents the number of points against the point cloud $x'$. 

Hausdorff distance is a common constraint term for generating 3D adversarial point clouds, which effectively reduces the outliers of the generated adversarial examples. Through this distance constraint, the generated adversarial examples can be made more imperceptible to a certain extent, making them more difficult to distinguish. The Hausdorff distance is calculated as follows:
\begin{equation}
	\begin{aligned}
		D_{\text {Hausdorff}}\left(x, x^{\prime}\right)=\max \left\{\max _{b \in x}\left\{\min _{a \in x^{\prime}}\left\|b-a\right\|_{2}\right\}\right. \text {, }\\
		\left.\max _{a \in x^{\prime}}\left\{\min _{b \in x}\left\|a-b\right\|_{2}\right\}\right\}. \\
	\end{aligned}
\end{equation}

In this paper, the local adversarial attack method using Chamfer distance and Hausdorff distance constraints can be expressed as follows:

\begin{equation}
	D(x, x') = 	D_{Chamfer}(x, x') + D_{Hausdorff}(x, x') .
\end{equation}

\textbf{Adversarial loss.}
Given a 3D network model $f$, input a 3D point cloud $x$ labeled $y$, the correct prediction result is $f(x)=y$. After attacking vulnerable regions of the input point cloud to generate an adversarial point cloud $x'$, our goal is to make the model produce the wrong output, i.e. $f(x')\neq y$. Therefore, we use an adversarial loss as follows to make the model fail to correctly identify adversarial point clouds.
\begin{equation}	
	l\left(x^{\prime}\right)=\max \left\{f_{y}\left(x^{\prime}\right)-\max _{y^{\prime} \neq y} f_{y^{\prime}}\left(x^{\prime}\right), 0\right\}.\\
\end{equation}
\noindent where $y$ is the class label of the original point cloud $x$. $f_{y}(x')$ indicates that the 3D network model recognizes the input $x'$ as class $y$. Here, the essence of adversarial loss is a penalty term. If the network model classifies the adversarial point cloud as the true class of the original point cloud, then $l(x')>0$, which requires a penalty. Otherwise, $l(x')=0$, and the adversarial loss will not work.

\begin{algorithm}[tb]
	\caption{Adaptive local adversarial attack}
	\label{alg:algorithm}
	\textbf{Input}: point cloud input $x$, label $y$, and model weights $\theta$; hyper-parameter $\lambda_{1}$,$\lambda_{2}$; number of iterations $T$.   \\
	\textbf{Output}: adversarial point cloud $x'$
	\begin{algorithmic}[1] 
		\STATE Initialize the adversarial point cloud $x'=x$
		\STATE Initialize perturbation $offset=0$
		\STATE Extract vulnerable regions $x=[a_1,a_2,\cdot\cdot\cdot,a_m]$
		\STATE Select the first $K$ regions for perturbation\\ $R = [p_1,\cdot\cdot\cdot,p_K]\in x$	
		\STATE Calculate the index $Region_{idx}$ of the first $K$ regions
		\FOR{$t=0$ to $T$ }
		\STATE Calculate the gradient of $x'$ \\ $grad=\bigtriangledown_{x}C_{Region}(x, x^{’})$
		\STATE Sign of the gradient: $sign(grad)$
		\STATE Calculate the proportion of disturbance size of the three-dimensional coordinate axis: \\ $Ration=[e_{x^*}, e_{y^*}, e_{z^*}]$
		\STATE Attack vulnerable region $R = [a_1,\cdot\cdot\cdot,a_K]$
		\STATE Update $x'= x +  Ration \cdot \epsilon \cdot \operatorname{sign}(grad) \cdot offset\cdot Region_{idx} $
		\ENDFOR
		\STATE \textbf{return} $x'$
	\end{algorithmic}
\end{algorithm}

\subsection{Adaptive gradient attack algorithm}
In order to generate more harmful adversarial examples, this paper designs an adaptive gradient attack algorithm for vulnerable regions.

Based on the vulnerability analysis of the 3D network model, it is only necessary to perturb the most vulnerable regions of the point cloud when generating the adversarial point cloud. So the salient values of all regions of the point cloud need to be sorted. Sort the salient values for different regions of the point cloud in descending order. Suppose the sorted point cloud is $x=(p_1,p_2,\cdot\cdot\cdot,p_k,p_m)$, where $\phi(p_1)>\phi(p_2)>\cdot\cdot\cdot>\phi(p_k)>\cdot\cdot\cdot>\phi(p_m)$. Then the adversarial attack method only needs to perturb the first $k$ regions $(p_1, p_2, ..., p_k)$ to produce adversarial examples.

The spatial location of 3D point cloud includes three directions: $x$, $y$, and $z$. The proposed adaptive gradient attack algorithm automatically assigns the disturbance size for each direction when disturbing vulnerable regions. First, the adaptive gradient attack algorithm calculates the gradient $grad=\bigtriangledown_{x}C_{Region}(x, x')$ of the adversarial point cloud for each iteration. Then, the calculation method of the ratio of the disturbance size in different coordinate axis directions is as follows. To avoid ambiguity, we denote the three axes of $x$, $y$, and $z$ as $x^*$, $y^*$, and $z^*$, respectively.
\begin{equation}	
	e_{x^*}= \dfrac{\mid grad_{x^*} \mid}{\mid grad_{x^*} \mid + \mid grad_{y^*} \mid + \mid grad_{z^*} \mid} ,
\end{equation}

\begin{equation}	
	e_{y^*}= \dfrac{\mid grad_{y^*} \mid}{\mid grad_{x^*} \mid + \mid grad_{y^*} \mid + \mid grad_{z^*} \mid} ,
\end{equation}

\begin{equation}	
	e_{z^*}= \dfrac{\mid grad_{z^*} \mid}{\mid grad_{x^*} \mid + \mid grad_{y^*} \mid + \mid grad_{z^*} \mid}.
\end{equation}

The $grad_{x^*}$, $grad_{y^*}$, and $grad_{z^*}$ in the above equation represent the gradient size of the adversarial point cloud on the three axes of $x$, $y$, and $z$, respectively. We denote the ratio of the disturbance size of the three-dimensional coordinate axis as $Ration=[e_{x^*}, e_{y^*}, e_{z^*}]$. After obtaining the ratio of the disturbance size of each coordinate axis in the three-dimensional space, the disturbance is automatically assigned to different coordinate axes in each iterative optimization process. The process of updating the adversarial point cloud once is expressed as follows:
\begin{equation}
	x^{\prime} \leftarrow x +  Ration \cdot \epsilon \cdot \operatorname{sign}(grad) \cdot offset\cdot Region_{idx}, \\
\end{equation}

\noindent where the parameter $\epsilon$ is the disturbance size and $offset$ is the disturbance to be optimized. $Region_{idx}$ represents the vulnerable regions that need to be disturbed. the $sign$ is a symbolic function, the specific calculation method is as follows:
\begin{equation}
	\operatorname{sign}(k)=\left\{\begin{array}{rl}
		1 & k>0 \\
		0 & k=0 \\
		-1 & k<0.
	\end{array}\right.
\end{equation}

\begin{table*}[!t]
	\centering
	\caption{The performance of different adversarial attack methods on ModelNet40 when the victim network is PointNet. Note: 'Success Rate' is higher as better; 'Chamfer Distance' is lower as better; 'Hausdorff Distance' is lower as better. '\# Points' represents the number of perturbed points, which is lower as better.}
	\label{table1}
	\begin{tabular}{c|c|c|c|c}
		\toprule  
		Methods   & Success Rate  $\uparrow$ & Chamfer Distance  $\downarrow$ & Hausdorff Distance  $\downarrow$ &   \# Points $\downarrow$ \\
		\midrule
		Jaeyeon Kim \cite{kim2021minimal}        & 89.38  & $1.55\times10^{-4}$  & $1.88\times10^{-2}$   & 36     \\
		Xiang et al \cite{xiang2019generating}     & 85.9   & $1.77\times10^{-4}$   & $2.38\times10^{-2}$   & 967     \\
		Adversarial sinK \cite{liu2020adversarial}   & 88.3   & $7.65\times10^{-3}$   & $1.92\times10^{-1}$   & 1024     \\
		Adversarial sticK \cite{liu2020adversarial}   & 83.7  & $4.93\times10^{-3}$   & $1.49\times10^{-1}$   & 210     \\
		Random selection \cite{wicker2019robustness}   & 55.56  & $7.47\times10^{-4}$   & $2.49\times10^{-3}$   & 413     \\
		Critical selectio \cite{wicker2019robustness}  & 18.99  & $1.15\times10^{-4}$   & $9.39\times10^{-3}$   & 50     \\
		Saliency map/critical frequency	\cite{zheng2019pointcloud}   & 63.15  & $5.72\times10^{-4}$  & $2.50\times10^{-3}$   & 303     \\
		Saliency map/low-score \cite{zheng2019pointcloud}            & 55.97  & $6.47\times10^{-4}$  & $2.50\times10^{-2}$   & 358     \\
		Saliency map/high-score \cite{zheng2019pointcloud}           & 58.39  & $7.52\times10^{-4}$  & $2.48\times10^{-3}$   & 424     \\
		
		\midrule
		AL-Adv \textbf{(Ours)}             & 92.92  & $2.36\times10^{-4}$  & $4.66\times10^{-2}$   & 40     \\
		
		\bottomrule
	\end{tabular}
\end{table*}

\subsection{Adaptive local adversarial attack method}
In this paper, the adaptive local adversarial attack method (AL-Adv) generates more harmful adversarial examples. The specific implementation process of the proposed AL-Adv is described as Algorithm \autoref{alg:algorithm}. The overall implementation process can be divided into two aspects. First, vulnerability analysis is performed on the 3D network model using Shapley value, and salient regions are extracted from the input point cloud. Second, the top $k$ vulnerable regions of the input point cloud are determined, and an adaptive gradient attack algorithm is applied to these vulnerable regions. Finally, the optimal adversarial point cloud is found in the iterative optimization process.

\section{Experiments}
\subsection{Experimental setup}
\textbf{Dataset.} This paper evaluates various adversarial attack methods on the ModelNet40 \cite{wu20153d} dataset. The ModelNet40 dataset is very popular for 3D network model classification and is one of the main evaluation datasets. This dataset contains 12311 CAD models of 40 classes. In specific experiments, the data used are point clouds obtained by uniform sampling from these CAD models. This paper follows the experimental setting in \cite{qi2017pointnet}, the training data uses a total of 9843 point clouds, and the test data contains 2468 point clouds.

\textbf{Implementation Details.}
In this paper, the experiments are only carried out on the 3D point cloud classification model. We adopt PointNet \cite{qi2017pointnet} as the attacked 3D network model. Following the previous setting \cite{qi2017pointnet,qi2017pointnet++}, the input point cloud of PointNet is 1024 points. In actual experiments, we select 25 samples of each class from the ModelNet40 test set for experiments. During model vulnerability analysis, we divided the input point cloud into 32 regions and calculated the salient value for each region. In the process of generating adversarial point clouds, we use Adam to optimize the perturbation, the learning rate of the optimizer is set to 0.01, the momentum is set to 0.9, and the perturbation size $\epsilon$ is set to 0.6. The parameter $\lambda_{1}$ is automatically adjusted during optimization using a binary search method and the $\lambda_{2}$ is set to 0.15. The number of salient regions attacked by the proposed method AL-Adv is five.

\textbf{Evaluation Metrics.}
The evaluation indicators of 3D adversarial attack methods mainly include attack success rate, Chamfer distance, Hausdorff distance, and the number of points manipulated when generating adversarial samples. This paper adopts these four metrics to measure different adversarial attack methods. Among them, the higher the attack success rate, the better, indicating that the generated adversarial examples have a greater threat to the network model. For the other three indicators, the smaller the better. Chamfer distance and Hausdorff distance measure how similar the adversarial example is to the original example. The number of points manipulated represents the cost of generating adversarial examples.

\subsection{Experimental results}
We conduct comparative experiments on PointNet. The comparison methods include the excellent 3D point cloud adversarial attack methods in recent years, which were proposed by Xiang et al. \cite{xiang2019generating}, Liu et al. \cite{liu2020adversarial}, Wicker et al. \cite{wicker2019robustness}, Zheng et al. \cite{zheng2019pointcloud}, Kim et al.\cite{kim2021minimal} respectively. According to the settings of Kim et al.\cite{kim2021minimal}, the method proposed by Xiang et al. \cite{xiang2019generating} was modified to untargeted attacks.The methods proposed by Liu et al. \cite{liu2020adversarial} are also reimplemented, denoted as Adversarial sink \cite{liu2020adversarial} and Adversarial stick \cite{liu2020adversarial}, respectively. Wicker et al. \cite{wicker2019robustness} generated adversarial examples by removing points, which was modified into perturbation, and implemented two attack methods Random selection \cite{wicker2019robustness} and Critical selection \cite{wicker2019robustness}. Kim et al.\cite{kim2021minimal} used the keypoint selection strategy of Zheng et al. \cite{zheng2019pointcloud}, but modified the point removal operation to a perturbation operation, and implemented three adversarial attack methods according to the rules of point removal, namely Saliency map/critical frequency \cite{zheng2019pointcloud}, Saliency map/low-score \cite{zheng2019pointcloud} ,and Saliency map/high-score \cite{zheng2019pointcloud}.

\autoref{table1} shows the comparison of various metrics for different adversarial attack methods. From the \autoref{table1}, the proposed method AL-Adv achieves the highest attack success rate, indicating that the AL-Adv  generates more harmful adversarial examples. At the same time, the number of points operated by the AL-Adv method is also relatively small, indicating that the cost of generating adversarial examples is low. In terms of distance metrics, both the Chamfer distance and the Hausdorff distance are relatively small, indicating that the adversarial examples generated by the method AL-Adv have good imperceptibility.

\begin{figure*}[!tb]
	\centering 
	\includegraphics[width= \linewidth]{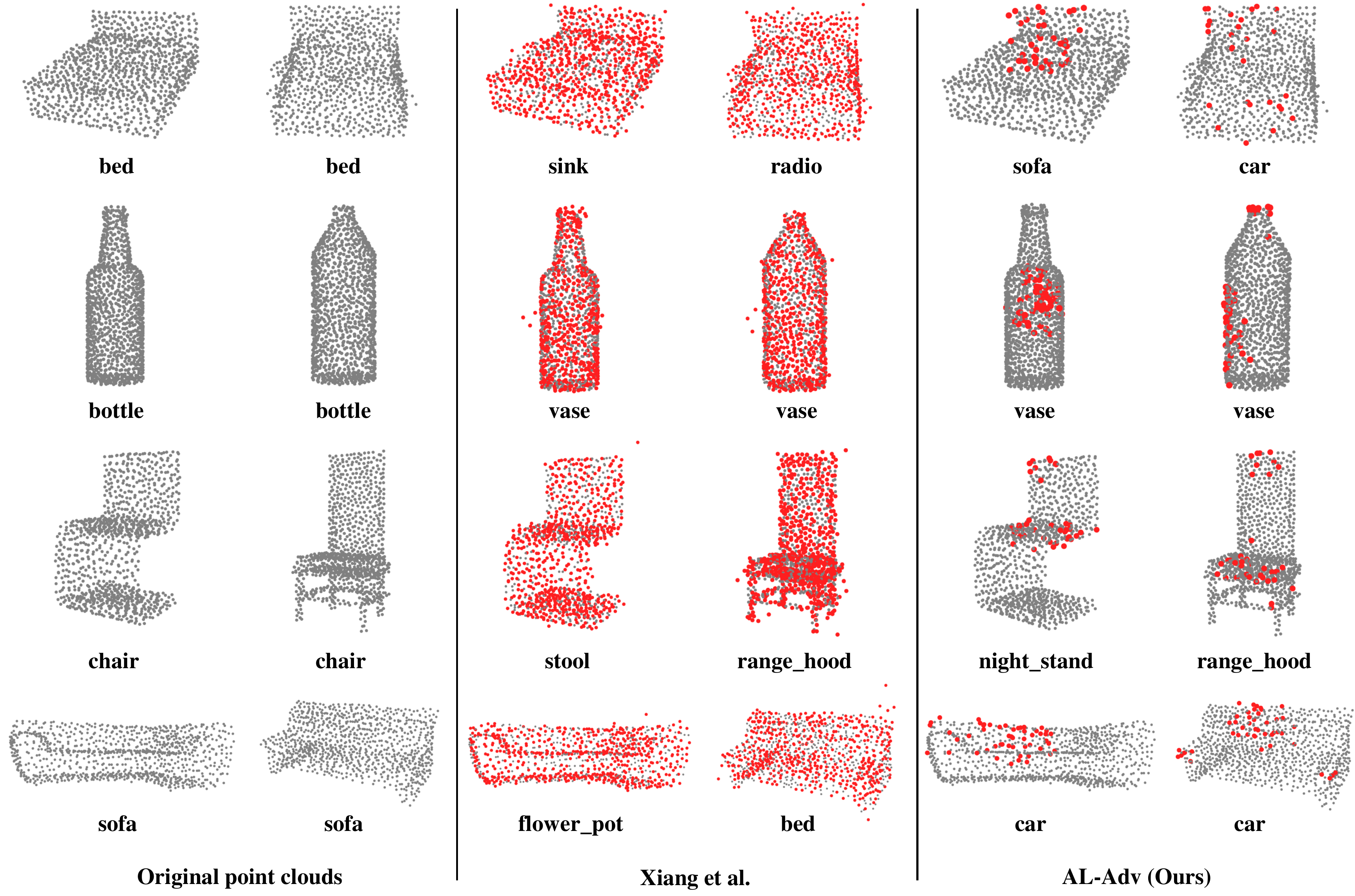}
	\caption{Adversarial point clouds generated by different adversarial attack methods on ModelNet40. The first and second columns are the original point clouds. The third and fourth columns are the adversarial point clouds generated by the attack method proposed by Xiang et al \cite{xiang2019generating}. The fifth and sixth columns are the adversarial point clouds generated by the proposed AL-Adv. The adversarial point clouds in the third and fifth columns come from the point clouds in the first column. The adversarial point clouds in the fourth and sixth columns come from the point clouds in the second column. The red points represent perturbed points when generating adversarial examples. The black points are the unperturbed points.}
	\label{fig:result}
\end{figure*}

In experiments, the proposed adversarial attack method (AL-Adv) targets local regions of point clouds and generates adversarial point clouds by attacking salient regions. All other comparison methods are global adversarial attack methods. Experimental results show that the proposed AL-Adv achieves the highest attack success rate of 92.92\% while only operating at 40 points. Although the method proposed by Kim et al. \cite{kim2021minimal} generates adversarial examples only by operating 36 points, the method achieves an attack success rate of 89.38\%. In addition, the "Adversarial sink" operated 1024 points to obtain an attack success rate of 88.3\%. Although the "Critical selection" method only uses 50 points to generate adversarial examples, the attack success rate is only 18.99\%, which shows that the quality of adversarial examples is low. In summary, the proposed local attack method (AL-Adv) achieves a higher attack success rate than the global attack method, and the AL-Adv achieves excellent performance overall.

\autoref{fig:result} shows the comparison of adversarial examples generated by the proposed method AL-Adv and the method proposed by Xiang et al. Compared with the original point cloud, the adversarial examples generated by Xiang et al. need to manipulate a large number of points, and are prone to produce obvious outliers that are easy to detect. However, the proposed AL-Adv operates fewer points when generating adversarial examples, and the generated adversarial examples have better imperceptibility. To highlight the difference in adversarial point clouds, we mark the perturbed points in red and increase the size of the red points.

In terms of performance evaluation, the attack success rate is an important indicator. The higher the attack success rate, the greater the threat to the network model caused by the generated adversarial examples. Therefore, the adversarial examples generated by the proposed AL-Adv tend to be more harmful to the network model. In addition, the cost and imperceptibility of the generated adversarial examples are also important metrics. To sum up, the proposed AL-Adv generates adversarial examples with good visual perception at a small cost, and the adversarial examples are more harmful.

\section{Discussion}
\textbf{Risks of AR Systems.} Due to the development of augmented reality (AR) technology, many industries have started the development and application of AR products. Therefore, the security issues of AR systems used in real life deserve attention. As an important support for AR technology, 3D recognition and tracking has an important impact on product safety. Point clouds are widely used in 3D object recognition and tracking due to their good data characteristics. For the development of AR, the application of point cloud data is also an important direction. In addition, deep network models are vulnerable to malicious attacks by adversarial examples, causing the model to fail to correctly recognize objects. Therefore, this paper focuses on the safety of AR systems on 3D point clouds and proposes a more harmful adversarial point cloud generation method AL-Adv.

\textbf{Advantages of Local regions Attacks.} Most of the current adversarial point cloud generation methods attack the entire point cloud, resulting in a large generation cost. Therefore, we pay more attention to the local regions of the point cloud. Analyzing the vulnerability of the 3D network model is more conducive to implementing adversarial attacks. Theoretically, we only need to attack the vulnerable part of the network model to achieve the adversarial attack. Therefore, the proposed method AL-Adv attacks only the most vulnerable region of the input point cloud for the 3D network model. The results demonstrate that AL-Adv generates adversarial point clouds at a small cost and that adversarial point clouds are more harmful. 

\textbf{Advantages of adaptive gradient attack.} Most of the existing gradient-based adversarial attack methods achieve the attacks in the gradient direction when perturbing the point cloud, which makes different dimensions of a point have the same size of perturbation. However, this is not reasonable, and there should be different perturbations size for different dimensions of a point. Therefore, this paper designs an adaptive gradient attack algorithm, so that each point in different dimensions automatically sets the perturbation size according to the gradient. Experiments show that the adaptive gradient attack algorithm achieves a higher attack success rate.

\section{Conclusion}
Adversarial objects in the physical world cause augmented reality (AR) systems to fail to perceive their surroundings. Therefore, adversarial objects have an important impact on the development of AR technology. In this paper, we propose an adaptive local adversarial attack method (AL-Adv) to generate 3D adversarial examples. First, we formulate the attack on local regions of point clouds. Then, we introduce the Shapley value to analyze the vulnerability of the 3D network model and extract the salient regions of the input point cloud. Finally, we design an adaptive gradient attack algorithm to attack these salient regions. The adaptive gradient attack algorithm adaptively assigns different disturbance sizes in different coordinate axis directions. Experimental results show that the proposed method AL-Adv has obtained a higher attack success rate than the global attack method, which shows that the adversarial examples generated by our method are more harmful to the 3D network model. Therefore, during the model training phase, the 3D network model learns adversarial examples to enhance model robustness, which is the basis for further improving the security of the AR system.


\clearpage

\bibliographystyle{abbrv-doi}

\bibliography{template}
\end{document}